\documentclass[11pt,a4paper]{article}
\usepackage[hyperref]{acl2019}
\usepackage{times}
\usepackage{latexsym}

\usepackage{url}
\usepackage{adjustbox}
\usepackage{multicol}
\usepackage{array,multirow,graphicx}
\usepackage{float}
\usepackage{subfigure}
\usepackage{hyperref}

\usepackage{tikz}
\usetikzlibrary{tikzmark, arrows.meta, decorations.pathreplacing}

\aclfinalcopy 

\title{The Effect of Translationese in Machine Translation Test Sets}

\author{Mike Zhang \\
  Information Science Programme\\
  University of Groningen \\
  The Netherlands \\
  \texttt{j.j.zhang.1@student.rug.nl} \\\And
  Antonio Toral\\
  Center for Language and Cognition\\
  University of Groningen\\
  The Netherlands\\
  \texttt{a.toral.ruiz@rug.nl}}

\date{}

\begin{document}
\maketitle
\begin{abstract}


The effect of translationese has been studied in the field of machine translation (MT), mostly with respect to training data.
We study in depth the effect of translationese on test data, using the test sets from the last three editions of WMT's news shared task, containing 17 translation directions.
We show evidence that (i) the use of translationese in test sets results in inflated human evaluation scores for MT systems; (ii) in some cases system rankings do change and (iii) the impact translationese has on a translation direction is inversely correlated to the translation quality attainable by state-of-the-art MT systems for that direction.
\end{abstract}

\section{Introduction}

Translated texts in a human language exhibit unique characteristics that set them apart from texts originally written in that language.
It is common then to refer to translated texts with the term {\it translationese}.
The characteristics of translationese can be grouped along the so-called universal features of translation or translation universals~\cite{baker1993corpus},
namely simplification, normalisation and explicitation. In addition to these three, interference is recognised as a fundamental law of translation~\cite{toury2012descriptive}: ``phenomena pertaining to the make-up of the source text tend to be transferred to the target text".
In a nutshell, compared to original texts, translations tend to be simpler, more standardised, and more explicit and they retain some characteristics that pertain to the source language.

The effect of translationese has been studied in machine translation (MT), mainly with respect to the training data, during the last decade.
Previous work has found that an MT system performs better when trained on parallel data whose source side is original and whose target side is translationese, rather than the opposite ~\cite{kurokawa2009automatic, lembersky2013effect}.

A recent paper has studied the effect of translationese on test sets~\cite{toral2018attaining}, in the context of assessing the claim of human parity made on Chinese-to-English WMT's 2017 test set~\cite{hassan2018achieving}.
The source side of this test set, as it is common in WMT~\citep{bojar2016findings, bojar2017findings, bojar2018findings}, was half original and half translationese.
It was found out that the translationese part was artificially easier to translate, which resulted in inflated scores for MT systems.

Noting that this finding was based on one test set for a single translation direction, we explore this topic in more depth, studying the effect of translationese in all the language pairs of the news shared task of WMT 2016 to 2018.
Our research questions (RQs) are the following:
\begin{itemize}

    \item RQ1. Does the use of translationese in the source side of MT test sets unfairly favour MT systems in general or is this just an artifact of the Chinese-to-English test set from WMT 2017?
    \item RQ2. If the answer to RQ1 is yes, does this effect of translationese have an impact on WMT's system rankings? In other words, would removing the part of the test set whose source side is translationese result in any change in the rankings?
    \item RQ3. If the answer to RQ1 is yes, would some language pairs be more affected than others? E.g. based on the level of the relatedness between the two languages involved.

\end{itemize}

The remainder of the paper will be organized as follows.
\autoref{s:related_work} provides an overview of previous work about the effect of translationese in MT.
Next, \autoref{s:data} describes the data sets used in our research.
This is followed by \autoref{s:rq1}, \autoref{s:rq2} and \autoref{s:rq3}, where we conduct the experiments for RQ1, RQ2 and RQ3, respectively.
Finally, \autoref{s:conclusions} outlines our conclusions and lines of future work.


\section{Related Work}\label{s:related_work}
There is previous research in the field of MT that has looked at the impact of translationese, mostly on training data, but there are works that have focused also on tuning and testing data sets.

The pioneering work on this topic by~\citet{kurokawa2009automatic} showed that French-to-English statistical MT systems trained on human translations from French to English (original source and translationese target, henceforth referred to as O$\rightarrow$T) outperformed systems trained on human translations in the opposite direction (i.e. translationese source and original target, henceforth referred to as T$\rightarrow$O).
These findings were corroborated by~\citet{lembersky2013effect}, who also adapted phrase tables to translationese, which resulted in further improvements.
~\citet{lembersky2012language} focused on the monolingual data used to train the language model of a statistical MT system and found that using translated texts led to better translation quality than relying on original texts.

~\citet{stymne2017effect} investigated the effect of translationese on tuning for statistical MT, using data from the WMT 2008--2013~\cite{bojar-EtAl:2013:WMT} for three language pairs. 
The results using O$\rightarrow$T and T$\rightarrow$O tuning texts were compared; the former led to a better length ratio and a better translation, in terms of automatic evaluation metrics. 

Finally,~\citet{toral2018attaining} investigated the effect of translationese
on the Chinese$\rightarrow$English (ZH$\rightarrow$EN) test set from WMT's 2017 news shared task.
They hypothesized that the sentences originally written in EN are easier to translate than those originally written in ZH, due to the simplification principle of translationese, namely that translated sentences tend to be simpler than their original counterparts~\cite{laviosa1998universals}. 
Two additional universal principles of translation, explicitation and normalisation, would also indicate that a ZH text originally written in EN would be easier to translate. 
In fact, they looked at a human translation and the translation by an MT system~\cite{hassan2018achieving} and observed that the human translation outperforms the MT system when the input text is written in the original language (ZH), but the difference between the two is not significant when the original language is translationese (ZH input originally written EN).
Therefore, they concluded that the use of translationese as the source language in test sets distorts the results in favour of MT systems.



\section{Data Sets}\label{s:data}

\begin{table*}[ht]
    \centering
    \resizebox{\linewidth}{!}{
    \begin{tabular}{l|rrr||rrr||rrr}
    
         Language Direction             & \multicolumn{3}{c||}{WMT16} & \multicolumn{3}{c||}{WMT17} & \multicolumn{3}{c}{WMT18} \\
         \hline\hline                               
                                        &   \# sys. & \# seg.      & \# assess.&  \# sys.  & \# seg.   & \# assess.& \# sys.   & \# seg.   & \# assess. \\\hline
        Chinese$\rightarrow$English     &           &              &           &   16      &  32,016   &  38,736   &   14      & 55,734    & 32,919     \\
        English$\rightarrow$Chinese     &           &              &           &   11      &  22,011   &  16,253   &   14      & 55,734    & 32,411    \\\hline
        Czech$\rightarrow$English       &  12       & 30,000       &  16,800   &   4       &  12,020   &  21,992   &   5       & 14,915    & 12,209    \\
        English$\rightarrow$Czech       &           &              &           &   14      &  42,070   &  32,564   &   5       & 14,915    & 10,080    \\\hline
        Estonian$\rightarrow$English    &           &              &           &           &           &           &   14      & 28,000    & 28,868    \\
        English$\rightarrow$Estonian    &           &              &           &           &           &           &   14      & 28,000    & 15,800    \\\hline
        Finnish$\rightarrow$English     &  9        & 63,040       &  30,080   &   6       &  18,012   &  27,545   &   9       & 27,000    & 18,868    \\
        English$\rightarrow$Finnish     &           &              &           &   12      &  36,024   &  8,289    &   12      & 36,000    & 9,995     \\\hline
        German$\rightarrow$English      &  10       & 68,800       &  33,760   &   11      &  33,044   &  36,189   &   16      & 47,968    & 48,469    \\
        English$\rightarrow$German      &           &              &           &   16      &  48,064   &  10,229   &   16      & 47,968    & 13,754    \\\hline
        Latvian$\rightarrow$English     &           &              &           &   9       &  18,009   &  30,321   &           &           &           \\
        English$\rightarrow$Latvian     &           &              &           &   17      &  34,017   &  6,882    &           &           &           \\\hline
        Romanian$\rightarrow$English    & 7         & 27,920       &  16,000   &           &           &           &           &           &           \\\hline
        Russian$\rightarrow$English     & 10        & 64,960       &  37,040   &   9       &  27,009   &  24,837   &  8        &  24,000   & 17,711    \\
        English$\rightarrow$Russian     &           &              &           &   9       &  27,009   &  25,798   &  9        &  27,000   & 27,977    \\\hline
        Turkish$\rightarrow$English     &  9        & 48,640       &  18,400   &   10      &  30,070   &  25,853   &  6        &  18,000   & 29,784    \\
        English$\rightarrow$Turkish     &           &              &           &   8       &  24,056   &  2,219    &  8        &  24,000   & 3,644     \\\hline    
    \end{tabular}
    }
    \caption{Datasets used in this study (DA scores from WMT16--18 news translation task). 
    Columns contain (from left to right) the number of submitted systems (\# sys.), total number of segments prior to quality control (\# seg.), and total number of assessments 
    human assessors carried out (\# assess.)} 

    \label{tab:data}
\end{table*}

We use the test data from WMT16, WMT17, and WMT18 news translation tasks (\textit{newstest2016, newstest2017, and newstest2018}) exclusively, because they provide results using the \textit{direct assessment} (DA) score~\cite{graham2013continuous, graham2014machine, graham2017can},
which is the metric we will use in our experiments. 
DA is a crowd-sourced human evaluation metric to determine MT quality. To elaborate, after participants submit their translations produced by their MT systems, a human evaluation campaign is run. 
This is to assess the translation quality of the systems, and to rank them accordingly. 
Human evaluation scores are provided via crowdsourcing and/or by participants, using
Appraise~\citep{federmann2012appraise}. 
Human assessors 
are asked to rate a given candidate translation by how adequately it expresses the meaning of the corresponding reference translation, thus avoiding the use of the source texts and therefore not requiring bilingual speakers. 
The rating is done on an analogue scale, which corresponds to an absolute 0-100 scale.

To prevent differences in scoring strategies of distinct human assessors, the human assessment scores for translations are standardized according to each individual human assessor's overall mean and standard deviation score, 
which is indicated as the $z$-score in WMT finding papers. 
Average standardized scores for individual segments belonging to a given system are then computed, before the final overall DA score for that system is computed as the average of its standardized segment scores. 

Finally, systems are ranked to produce the shared task results.
There is of course the possibility that some systems score similarly in the shared task. 
If that is the case, those systems are clustered together. 
Specifically, clusters are determined by grouping systems together, and comparing the scores they obtained. According to the Wilcoxon rank-sum test, if systems do not significantly outperform others, they are in the same cluster, the opposite is the case if they do outperform each other~\citep{bojar2016findings,bojar2017findings,bojar2018findings}.
\autoref{tab:data} provides an overview of the number of systems, segments, and assessments in the previously mentioned editions of WMT for all available language directions.
These are the datasets that we use in this work.


\section{Effect of Translationese on Direct Assessment Scores}\label{s:rq1}

The test sets used by~\citet{bojar2016findings,bojar2017findings,bojar2018findings} are bilingual, thus having two sides: source text and reference translation. 
The source is written in the language that is to be translated from (original language), while the reference is written in the language into which the source text is to be translated (target language). 
In all the test sets used in our experiments English is one of the two languages involved, being either the source or the target.

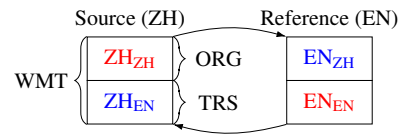
\begin{figure}[ht]
\centering
\scalebox{.75}{
\begin{tikzpicture}[minimum size = 2em, minimum width = 1.5cm, align=center]
	\node[rectangle, draw](tl) {\textcolor{red}{ZH\textsubscript{ZH}}};
	\node[rectangle, draw, right of = tl, node distance=3.5cm](tr) {\textcolor{blue}{EN\textsubscript{ZH}}};
	\node[rectangle, draw, below of = tl, node distance = 2em](bl) {\textcolor{blue}{ZH\textsubscript{EN}}};
	\node[rectangle, draw, below of = tr, node distance = 2em](br) {\textcolor{red}{EN\textsubscript{EN}}};
	\draw[decorate, line width=.4pt, decoration={amplitude=6pt,brace,mirror}] (tl.north west) -- (bl.south west) node[left of = tl, anchor = east, node distance=.8cm, yshift=-1em] {WMT};
	\draw[decorate, line width=.4pt, decoration={amplitude=6pt,brace}] (tl.north east) -- (tl.south east) node[right of = tl, anchor = west, node distance=.8cm] {ORG};
	\draw[decorate, line width=.4pt, decoration={amplitude=6pt,brace}] (bl.north east) -- (bl.south east) node[right of = bl, anchor = west, node distance=.8cm] {TRS};
	\node[above of = tl, anchor = center, node distance = .7cm] {Source (ZH)};
	\node[above of = tr, anchor = center, node distance = .7cm] {Reference (EN)};
	\draw[{Latex[width=3pt]}-] (bl.south east) to [out = -15, in = -165] (br.south west);
	\draw[-{Latex[width=3pt]}] (tl.north east) to [out = 15, in = 165] (tr.north west);
\end{tikzpicture}}
\caption{Example of a WMT test set for English (EN) $\rightarrow$ Chinese (ZH) translation direction, where English is translated into Chinese, and Chinese into English. Indicated as a subscript is which the original language was, red means original language and blue translationese.}
\label{fig:dist}
\end{figure}

Taking as an example of WMT test set the one for Chinese-to-English from 2017, this contains 2,001 sentence pairs.
Out of these, 1,000 sentences were originally written in Chinese and translated by a human translator into English, hence the target text is translationese.
The other half consists of 1,001 sentences that were originally written in English and translated by a human translator into Chinese, hence the source text is translationese in this subset.
A graphical depiction of this can be found in Figure~\ref{fig:dist}.
The advantage of this procedure is that the same test set can be used for the English-to-Chinese direction, thus reducing the costs involved in creating test sets in half.

\begin{table*}[ht]
    \centering
    \begin{tabular}{l|lll||lll||lll}
        Language Direction              & \multicolumn{3}{c||}{WMT16} & \multicolumn{3}{c||}{WMT17} & \multicolumn{3}{c}{WMT18} \\
        \hline\hline                               
                                        & WMT   & ORG   & TRS   &   WMT     & ORG   & TRS   & WMT   & ORG   & TRS   \\\hline
        Chinese$\rightarrow$English     &           &           &           & 73.2          & {-1.5}    & {+3.9}    & 78.8      & {-1.3}    & {+2.0}    \\
        English$\rightarrow$Chinese     &           &           &           & 73.2          & {-4.1}    & {+5.0}    & 80.7      & {-4.0}    & {+2.3}    \\\hline
        Czech$\rightarrow$English       & 75.4      & {-5.8}    & {+5.7}    & 74.6          & {-4.3}    & {+4.2}    & 71.8      & {-1.6}    & {+1.6}    \\
        English$\rightarrow$Czech       &           &           &           & 62.0          & {-5.8}    & {+7.4}    & 67.2      & {-6.6}    & {+7.2}    \\\hline
        Estonian$\rightarrow$English    &           &           &           &               &           &           & 73.3      & {-4.0}    & {+4.0}    \\
        English$\rightarrow$Estonian    &           &           &           &               &           &           & 64.9      & {-4.1}    & {+3.9}    \\\hline
        Finnish$\rightarrow$English     & 66.9      & {-3.2}    & {+3.0}    & 73.8          & {-2.1}    & {+2.2}    & 75.2      & {-2.4}    & {+2.3}    \\
        English$\rightarrow$Finnish     &           &           &           & 59.6          & {-5.1}    & {+5.6}    & 64.7      & {-7.7}    & {+8.0}    \\\hline
        German$\rightarrow$English      & 75.8      & {-4.1}    & {+4.1}    & 78.2          & {-2.4}    & {+2.2}    & 79.9      & {-3.8}    & {+4.3}    \\
        English$\rightarrow$German      &           &           &           & 72.9          & {-5.1}    & {+4.4}    & 85.5      & {-1.9}    & {+1.9}    \\\hline
        Latvian$\rightarrow$English     &           &           &           & 76.2          & {-0.4}    & {+0.6}    &           &           &           \\
        English$\rightarrow$Latvian     &           &           &           & 54.4          & {-11.2}   & {+11.7}   &           &           &           \\\hline
        Romanian$\rightarrow$English    & 73.9      & {-0.4}    & {+0.5}    &               &           &           &           &           &           \\\hline
        Russian$\rightarrow$English     & 74.2      & {-1.2}    & {+1.8}    & 82.0          & {-0.7}    & {+0.6}    & 81.0      & {-0.1}    &  0.0     \\
        English$\rightarrow$Russian     &           &           &           & 75.4          & {-5.8}    & {+5.8}    & 72.0      & {-7.4}    & {+7.4}    \\\hline
        Turkish$\rightarrow$English     & 57.1      & {-1.6}    & {+1.6}    & 68.8          & {-3.8}    & {+3.9}    & 74.3      & {-3.2}    & {+3.9}    \\
        English$\rightarrow$Turkish     &           &           &           & 53.4          & {-13.4}   & {+11.8}   & 66.3      & {-4.1}    & {+5.5}    \\\hline
    \end{tabular}
    \caption{DA scores for the best MT system for each translation direction of WMT's 2016--2018 news translation shared task. Columns ORG and TRS show the absolute difference of the DA scores in those subsets compared to the whole test set (WMT).}
    \label{tab:overall}
\end{table*}


Source and reference files contain documents, each of which is provided with a label indicating in which language it was originally written. 
In our experiments we compute the DA scores for each test set (i) on the whole test set, which corresponds to the results reported in WMT, (ii) on the subset for which the source text was originally written in the source language (referred to as ORG in our experiments) and (iii) on the remaining subset, for which the source text was originally written in the target language, and is thus translationese (referred to as TRS in our experiments).


\autoref{tab:overall} shows the absolute difference in DA score for the ORG and TRS subsets, taking the whole test set (WMT) as starting point for the comparison. 
We observe a clear and common trend: using original input results in a lower DA score, while using translationese input increases the DA score.
This trend is consistent for all the 17 translation directions considered and for all the 3 years of WMT studied,
thus providing enough evidence to answer RQ1: the use of translationese as input of test sets results in higher DA scores for MT systems.



\begin{table*}[htbp]
	\centering
	\resizebox{\linewidth}{!}{
			\begin{tabular}{l|ccc|c||c|ccc|r}
				& \multicolumn{3}{c|}{\textbf{With Ties}}& \multicolumn{2}{c|}{\multirow{2}{*}{\textbf{Mean}}} &\multicolumn{3}{c|}{\textbf{Without Ties}}\\
				\cline{1-4} \cline{7-10}
				Language Direction   & WMT16   &    WMT17 &   WMT18   &  \multicolumn{2}{c|}{}  &   WMT16 &   WMT17 &   WMT18 & Language Direction\\
				\hline
				\hline
				
				Romanian   $\rightarrow$   English$\dagger$&  1.000* &       - &       - & 1.000 & 1.000  & 1.000* &       - &       -    & Romanian       $\rightarrow$   English $\dagger$    \\\hline
				Turkish    $\rightarrow$   English       &  0.983* &  0.948* &  1.000* & 0.977 & 1.000  & 1.000* &  1.000* &  1.000*    &  Czech          $\rightarrow$   English          \\\hline
				Finnish    $\rightarrow$   English       &  0.943* &  0.966* &  1.000* & 0.970 & 0.978  &      - &       - &  0.978*    &  English        $\rightarrow$   Estonian  $\dagger$\\\hline
				Czech      $\rightarrow$   English       &  0.929* &  1.000* &  0.949* & 0.959 & 0.956  &      - &       - &  0.956*    &  Estonian       $\rightarrow$   English   $\dagger$\\\hline
				German     $\rightarrow$   English       &  0.979* &  0.939* &  0.906* & 0.941 & 0.944  &      - &  0.944* &       -    &  Latvian        $\rightarrow$   English   $\dagger$\\\hline
				English    $\rightarrow$   Czech         &       - &  0.904* &  0.949* & 0.927 & 0.929  &      - &  0.929* &  0.929*    &  English        $\rightarrow$   Turkish          \\\hline
				Latvian    $\rightarrow$   English$\dagger$&       - &  0.921* &       - & 0.921 & 0.917  &      - &  0.889* &  0.944*    &  English        $\rightarrow$   Russian            \\\hline
				English    $\rightarrow$   Finnish       &       - &  0.868* &  0.968* & 0.918 & 0.898  &      - &  0.927* &  0.868*    & English        $\rightarrow$   Chinese            \\\hline
				English    $\rightarrow$   Russian       &       - &  0.873* &  0.935* & 0.904 & 0.882  &      - &  0.882* &       -    &  English        $\rightarrow$   Latvian    $\dagger$ \\\hline
				Chinese    $\rightarrow$   English       &       - &  0.923* &  0.882* & 0.903 & 0.869  & 0.733* &  0.944* &  0.929*    & Russian        $\rightarrow$   English            \\\hline
				English    $\rightarrow$   German        &       - &  0.863* &  0.856* & 0.860 & 0.852  & 1.000* &  1.000* &  0.556*    & Finnish        $\rightarrow$   English            \\\hline
				English    $\rightarrow$   Estonian$\dagger$&       - &       - &  0.845* & 0.845 & 0.848  & 0.833* &  0.911* &  0.800*    & Turkish        $\rightarrow$   English            \\\hline
				Estonian   $\rightarrow$   English$\dagger$&       - &       - &  0.830* & 0.830 & 0.784  &      - &  0.633* &  0.934*    & Chinese        $\rightarrow$   English            \\\hline
				English    $\rightarrow$   Chinese       &       - &  0.847* &  0.789* & 0.818 & 0.726  &      - &  0.451* &  1.000*    & English        $\rightarrow$   Czech              \\\hline
				English    $\rightarrow$   Turkish       &       - &  0.890* &  0.734* & 0.812 & 0.713  & 0.911* &   0.345 &  0.883*    & German         $\rightarrow$   English          \\\hline
				Russian    $\rightarrow$   English       &   0.557 &  0.845* &  0.890* & 0.764 & 0.675  &      - &  0.817* &  0.533*    &  English        $\rightarrow$   German             \\\hline
				English    $\rightarrow$   Latvian $\dagger$&       - &  0.718* &       - & 0.718 & 0.637  &      - &  0.970* &  0.303     & English        $\rightarrow$   Finnish            \\\hline
			\end{tabular}
	}
	\caption{Kendall's $\tau$ coefficient for each translation direction and year. The coefficient is obtained by comparing WMT's ranking with the ranking if only original language is used as input (subset ORG), with and without ties. A (*) indicates the significance level at p-level p$\leq$0.05. Furthermore, language directions are sorted by the computed mean Kendall's $\tau$. A $\dagger$ indicates that the mean is computed over one year.} \label{tab:kendall}
\end{table*}


\begin{table*}[p]
\resizebox{\linewidth}{!}{
\begin{tabular}{lclrr|cclrr|cclrr}
\multicolumn{15}{c}{\textbf{Chinese$\rightarrow$English}}\\\\
& \# &           SYSTEM   &   RAW.WMT &   Z.WMT & \# &$\uparrow\downarrow$    &             SYSTEM & RAW.ORG & Z.ORG     & \# &   $\uparrow\downarrow$&           SYSTEM     & RAW.TRS &    Z.TRS \\ \hline \hline
&   1& SogouKnowing-nmt   &      73.2 &   0.209 &   1& $2^{\uparrow}$         & xmunmt             &    71.7 &     0.167 &   1&         $1^{\uparrow}$& uedin-nmt            &    77.1 &    0.316 \\
&    & uedin-nmt          &      73.8 &   0.208 &    & $1^{\downarrow}$       & SogouKnowing-nmt   &    71.9 &     0.161 &    &       $1^{\downarrow}$& SogouKnowing-nmt     &    74.4 &    0.257 \\
&    & xmunmt             &      72.3 &   0.184 &    & $1^{\downarrow}$       & uedin-nmt          &    70.5 &     0.101 &   3&         $2^{\uparrow}$& online-A             &    73.6 &    0.208 \\
&   4& online-B           &      69.9 &   0.113 &    & $-$                    & online-B           &    68.7 &     0.081 &    &       $1^{\downarrow}$& xmunmt               &    72.9 &    0.202 \\
&    & online-A           &      70.4 &   0.109 &    & $1^{\uparrow}$         & NRC                &    69.1 &     0.064 &   5&       $1^{\downarrow}$& online-B             &    71.1 &    0.145 \\
&    & NRC                &      69.8 &   0.079 &   6& $1^{\downarrow}$       & online-A           &    67.4 &     0.012 &    &         $1^{\uparrow}$& jhu-nmt              &    70.0 &    0.110 \\\parbox[t]{3mm}{\multirow{1}{*}{\rotatebox[origin=c]{90}{\textbf{wmt17}}}} 
&   7& jhu-nmt            &      67.9 &   0.023 &   7& $-$                    & jhu-nmt            &    65.8 &    -0.062 &    &       $1^{\downarrow}$& NRC                  &    70.4 &    0.093 \\
&   8& afrl-mitll-opennmt &      66.9 &  -0.016 &    & $1^{\uparrow}$         & CASICT-cons        &    65.4 &    -0.087 &    &                    $-$& afrl-mitll-opennmt   &    69.2 &    0.063 \\
&    & CASICT-cons        &      67.1 &  -0.026 &    & $1^{\downarrow}$       & afrl-mitll-opennmt &    64.5 &    -0.095 &    &                    $-$& CASICT-cons          &    68.9 &    0.036 \\
&    & ROCMT              &      65.4 &  -0.058 &    & $-$                    & ROCMT              &    63.4 &    -0.108 &    &                    $-$& ROCMT                &    67.4 &   -0.006 \\
&  11& Oregon-State-Uni-S &      64.3 &  -0.107 &    & $-$                    & Oregon-State-Uni-S &    62.7 &    -0.162 &    &                    $-$& Oregon-State-Uni-S   &    65.9 &   -0.054 \\
&  12& PROMT-SMT          &      61.7 &  -0.209 &  12& $3^{\uparrow}$         & online-F           &    60.0 &    -0.261 &  12&                    $-$& PROMT-SMT            &    64.0 &   -0.137 \\
&    & NMT-Ave-Multi-Cs   &      61.2 &  -0.265 &    & $1^{\downarrow}$       & PROMT-SMT          &    59.4 &    -0.282 &    &                    $-$& NMT-Ave-Multi-Cs     &    63.3 &   -0.193 \\
&    & UU-HNMT            &      60.0 &  -0.276 &    & $-$                    & UU-HNMT            &    58.8 &    -0.301 &  14&         $2^{\uparrow}$& online-G             &    61.1 &   -0.245 \\
&    & online-F           &      59.6 &  -0.279 &    & $2^{\downarrow}$       & NMT-Ave-Multi-Cs   &    59.2 &    -0.337 &    &       $1^{\downarrow}$& UU-HNMT              &    61.1 &   -0.251 \\
&    & online-G           &      59.3 &  -0.305 &    & $-$                    & online-G           &    57.4 &    -0.363 &    &       $1^{\downarrow}$& online-F             &    59.2 &   -0.296 \\
\hline
&   1& NiuTrans           &      78.8 &   0.140 &   1& $-$                     & NiuTrans           &    77.5 &     0.091 &   1&        $8^{\uparrow}$& UMD                  &    80.8 &    0.239 \\
&    & online-B           &      77.7 &   0.111 &    & $-$                     & online-B           &    77.4 &     0.089 &    &        $6^{\uparrow}$& NICT                 &    80.5 &    0.232 \\
&    & UCAM               &      77.9 &   0.109 &    & $2^{\uparrow}$          & Tencent-ensemble   &    77.0 &     0.067 &    &      $2^{\downarrow}$& NiuTrans             &    81.1 &    0.222 \\
&    & Unisound-A         &      78.0 &   0.108 &    & $1^{\downarrow}$        & UCAM               &    76.3 &     0.048 &    &                   $-$& Unisound-A           &    80.9 &    0.222 \\
&    & Tencent-ensemble   &      77.5 &   0.099 &    & $1^{\downarrow}$        & Unisound-A         &    76.4 &     0.041 &    &        $2^{\uparrow}$& Li-Muze              &    80.7 &    0.214 \\
&    & Unisound-B         &      77.5 &   0.094 &    & $-$                     & Unisound-B         &    75.8 &     0.029 &    &      $3^{\downarrow}$& UCAM                 &    80.5 &    0.211 \\
\parbox[t]{3mm}{\multirow{1}{*}{\rotatebox[origin=c]{90}{\textbf{wmt18}}}}
&    & Li-Muze            &      77.9 &   0.091 &    & $-$                     & Li-Muze            &    76.2 &     0.016 &    &      $1^{\downarrow}$& Unisound-B           &    80.5 &    0.206 \\
&    & NICT               &      77.0 &   0.089 &    & $-$                     & NICT               &    75.0 &     0.004 &    &        $3^{\uparrow}$& uedin                &    79.6 &    0.180 \\
&    & UMD                &      76.7 &   0.078 &    & $-$                     & UMD                &    74.3 &    -0.021 &    &      $4^{\downarrow}$& Tencent-ensemble     &    78.1 &    0.149 \\
&  10& online-Y           &      75.0 &  -0.005 &    & $-$                     & online-Y           &    73.8 &    -0.047 &    &      $8^{\downarrow}$& online-B             &    78.1 &    0.147 \\
&    & uedin              &      74.5 &  -0.017 &  11& $-$                     & uedin              &    71.5 &    -0.137 &  11&        $1^{\uparrow}$& online-A             &    77.1 &    0.068 \\
&  12& online-A           &      73.6 &  -0.061 &    & $-$                     & online-A           &    71.4 &    -0.140 &    &      $2^{\downarrow}$& online-Y             &    76.8 &    0.061 \\
&  13& online-G           &      65.9 &  -0.327 &  13& $1^{\uparrow}$          & online-F           &    65.2 &    -0.353 &  13&                   $-$& online-G             &    67.8 &   -0.262 \\
&  14& online-F           &      64.4 &  -0.377 &    & $1^{\downarrow}$        & online-G           &    64.9 &    -0.364 &  14&                   $-$& online-F             &    63.1 &   -0.417 \\
\hline
\end{tabular}}
\caption{Results of the Chinese$\rightarrow$English language direction with WMT, ORG, and TRS input. Systems are ordered by standardized mean DA score. If a system does not contain a rank, this means that it shares the same cluster as the system above it. Clusters are obtained according to Wilcoxon rank-sum test at p-level p $\leq$ 0.05. Indicated in the [$\uparrow\downarrow$] column are the changes in absolute ranking (i.e. how many positions a system goes up or down).} \label{tab:zhen}
\end{table*}

\section{Effect of Translationese on Rankings}\label{s:rq2}

We compute Kendall's $\tau$ to give an overview of to what degree rankings change for each translation direction. 
The $\tau$ coefficient is obtained by comparing WMT rankings to the resulting rankings if only the ORG subset is used as input.
Since systems can share the same cluster, and thus the same ranking, we compute Kendall's $\tau$ both with and without ties. 
With ties, all systems in the same cluster are considered to occupy the same rank, hence the correlation with ties is sensitive only to changes that go beyond clusters. E.g. if a system moves from the second cluster to the first one.
In contrast, without ties all the ranking changes are considered, even if a system changes position but remains within the same cluster.

\autoref{tab:kendall} shows the Kendall's $\tau$ correlations for all translation directions between the rankings on the whole test set (WMT) and on the ORG subset.
We do see that some of the translation directions have a $\tau$ coefficient of 1, which means that the agreement between the two rankings is perfect, i.e. the rankings in WMT and ORG are exactly the same.
However, we observe that there were few systems submitted to such translation directions (e.g. $\tau=1$ for Romanian$\rightarrow$English in 2017, for which 7 systems were submitted, see \autoref{tab:data}). 
Apart from those, other language directions show that there are at least slight rank changes between the WMT rankings and ORG rankings.
Looking at the low ranked translation directions, we observe that some are close to a $\tau$ coefficient of 0, especially in correlations without ties, such as German$\rightarrow$English in WMT 2017 ($\tau=0.345$). 
This means that some rankings have only a weak correlation.

Probably related to the differences in DA scores between WMT and ORG (RQ1), we also find that systems' rankings change for most language pairs 
when comparing WMT and ORG rankings. 
We see that there is no perfect correlation between rankings, apart from a few language directions for which only a few systems were submitted. 
This indicates that the rankings do change to a certain degree. Computing Kendall's $\tau$ with ties results in higher correlation coefficients than without ties, implying that systems do shift, but tend to stay in the same cluster they occupied in the WMT ranking. 
In some editions of WMT, the rankings for certain language pairs change considerably. 
The biggest change in terms of ranking takes place for PROMT's rule-based system RU$\rightarrow$EN for WMT16. This system advances four positions in the ranking when only original source text is considered, going from rank 5 to rank 1 (although tied with several other systems).
It is worth noting that while the DA score for the majority of systems decreases when using original source text, the opposite happens for PROMT's system.


Thus far we have looked at a single result per translation direction and year, based on the best system in \autoref{tab:overall}, and on the correlation between systems in \autoref{tab:kendall}.
Now we zoom in on a translation direction: Chinese$\rightarrow$English.
\autoref{tab:zhen} shows how DA scores change between the whole test set (WMT) and the subsets ORG and TRS, both in terms of raw and standarized scores.
In addition, the table depicts how many positions a system goes up or down in the ranking. 

In the table we observe consistently that the DA score for ORG input is lower than that for WMT, while that for TRS is higher than that for WMT. 
It is also worth noting that most top scoring systems change in rankings, and that system clusters shift.
Due to limited space we provide equivalent tables to \autoref{tab:zhen} for the remaining 16 translation directions as an appendix.


\section{Effect of Translationese on Different Language Pairs}\label{s:rq3}

We aim to find out not only whether translationese has an effect on test sets (RQ1 and RQ2), but also to study whether some language pairs are more affected than others (RQ3).
Two hypotheses in this regard are as follows:
(i) the degree of translationese's impact has to do with the translation quality attainable for a translation direction, as represented by the DA score of the best MT system submitted;
(ii) the degree of translationese's impact has to do with how related are the two languages involved.

In order to test the second hypothesis, the degree of similarity between languages has to be quantified. 
We make use of the lang2vec tool~\citep{littell2017uriel} using the URIEL Typological Database~\citep{littell2016uriel} to compute the similarity between pairs of languages. 
Similar to the approach of~\citet{berzak2017predicting}, all the 103 available morphosyntactic features in URIEL are obtained; these are derived from the World Atlas of Language Structures (WALS)~\cite{dryer2013wals}, Syntactic Structures of the World’s Languages (SSWL)~\cite{collins2009syntactic} and Ethnologue~\cite{lewis2009ethnologue}. 
Missing feature values are filled with a prediction from a $k$-nearest neighbors classifier.
We also extract URIEL's 3,718 language family features derived from Glottolog~\cite{hammarstrom2019glottolog}. 
Each of these features represents membership in a branch of Glottolog's world language tree. 
Truncating features with the same value for all the languages present in our study, 87 features remain, consisting of 60 syntactic features and 27 family tree features. 
We then measure the level of relatedness between two languages using the linguistic similarity (LS) by~\citet{berzak2017predicting} (\autoref{eq:ls}), i.e. the cosine similarity between the URIEL feature vectors for two languages $v_y$ and $v_y^{\prime}$.

\begin{equation}
L S_{y, y^{\prime}}=\frac{v_{y} \cdot v_{y^{\prime}}}{\left\|v_{y}\right\|\left\|v_{y^{\prime}}\right\|}
\label{eq:ls}
\end{equation}

Together with the LS for a language direction, we take the best system of the most recent year in our data set, WMT18, for that language direction. 
The motivation behind is that a top performing system from the most recent campaign should be representative of the current state-of-the-art in machine translation for the translation direction it was submitted to.

To look into the effect of translationese across different language pairs, we present two approaches, following the hypotheses put forward at the beginning of this section: (i) compare the DA score of the best system for each translation direction on subset ORG to the relative or absolute difference in DA score for that system between subset ORG and the whole set (WMT);
(ii) compare the LS of the two languages in each translation direction to the relative or absolute difference in DA scores for the best system between subset ORG and the whole set (WMT);


\begin{figure*}[ht]
	\includegraphics[width=.5\linewidth,page=1]{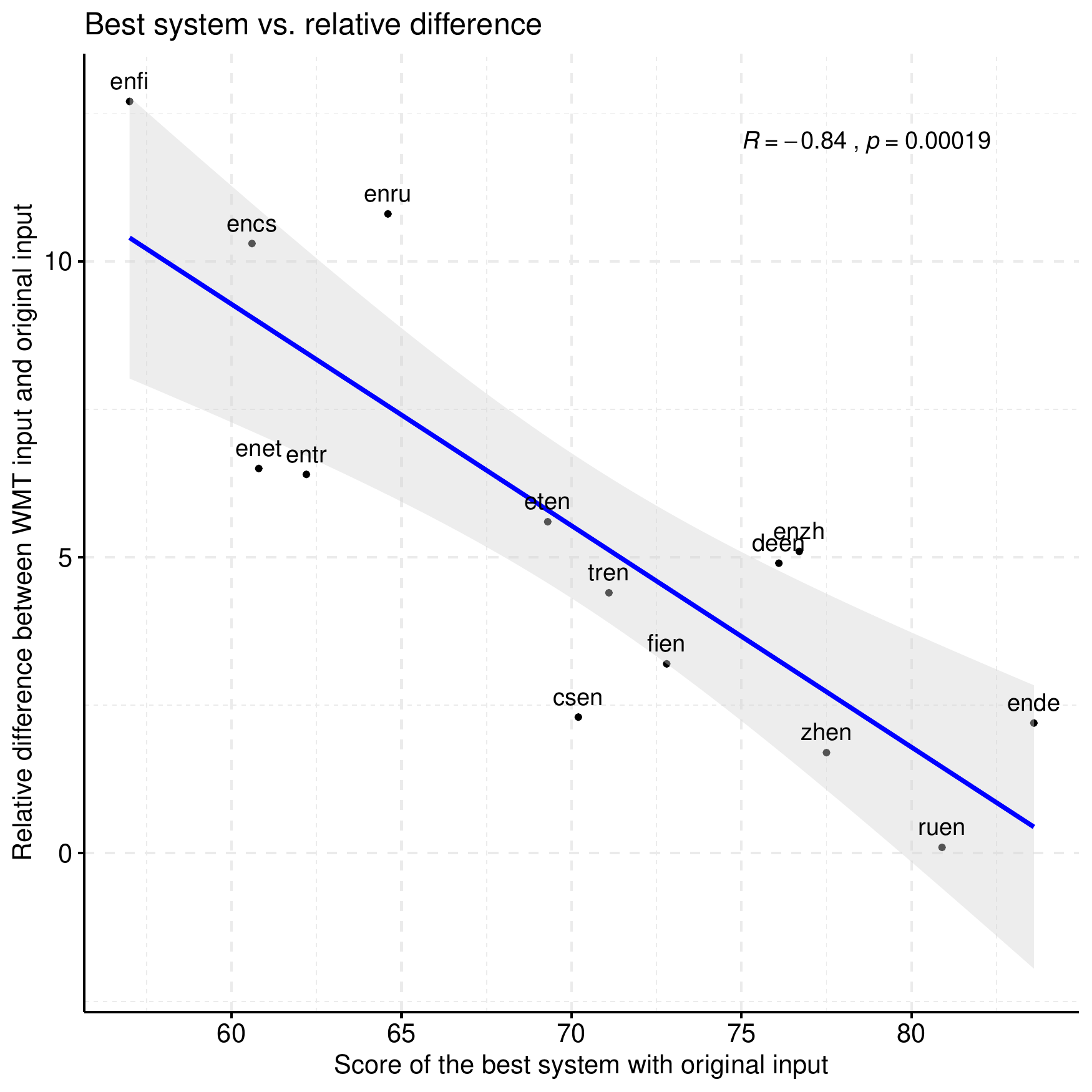}
    \includegraphics[width=.5\linewidth,page=2]{Rplots.pdf}
    \caption{Pearson correlation between the DA scores of the best system for each translation direction at WMT18 and the relative (left) and absolute (right) difference in DA score (\%) of comparing WMT input and ORG input. The languages are abbreviated into ISO 639-1 codes~\citep{byrum1999iso}.}
    \label{fig:org.vs.abs/rel}
\end{figure*}
\begin{figure*}[!t]
    \includegraphics[width=.5\linewidth,page=3]{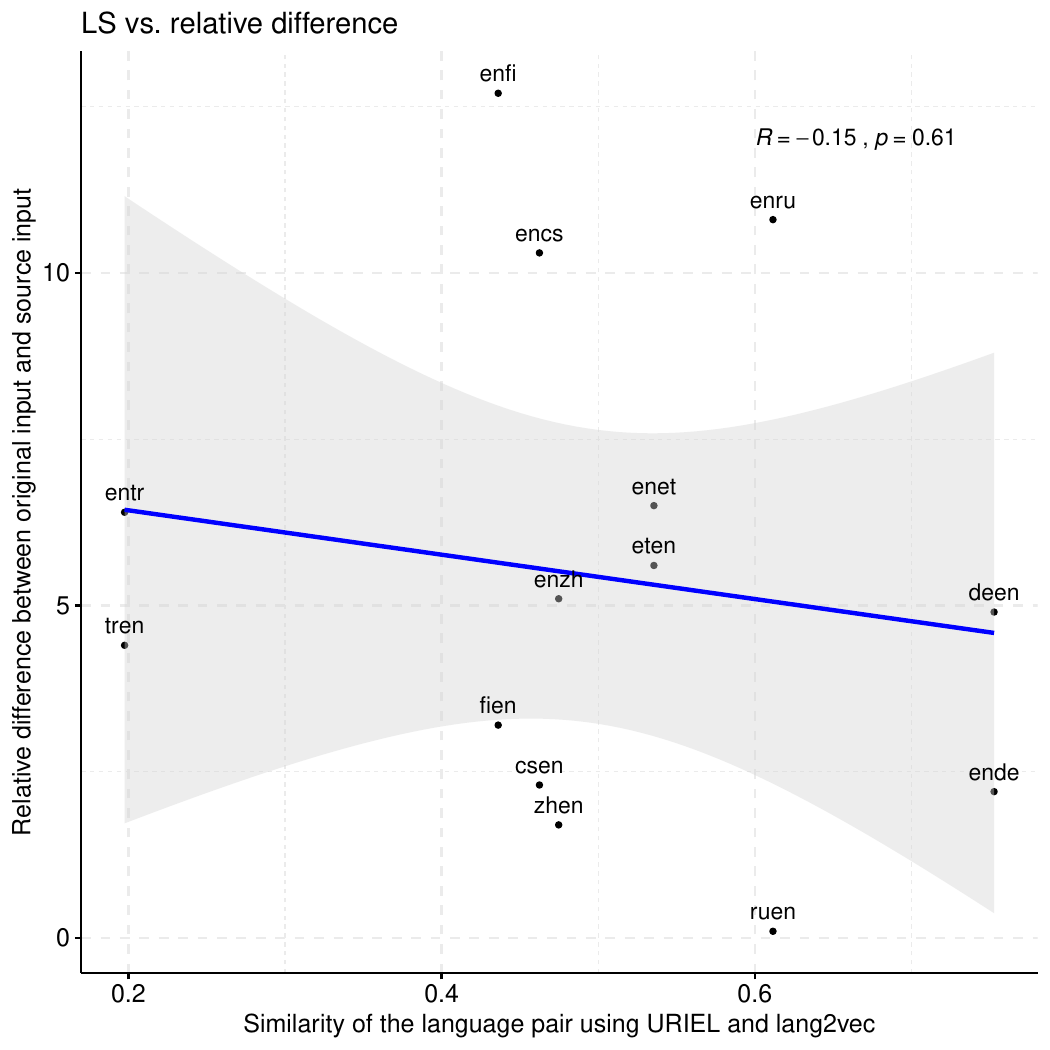}
    \includegraphics[width=.5\linewidth,page=4]{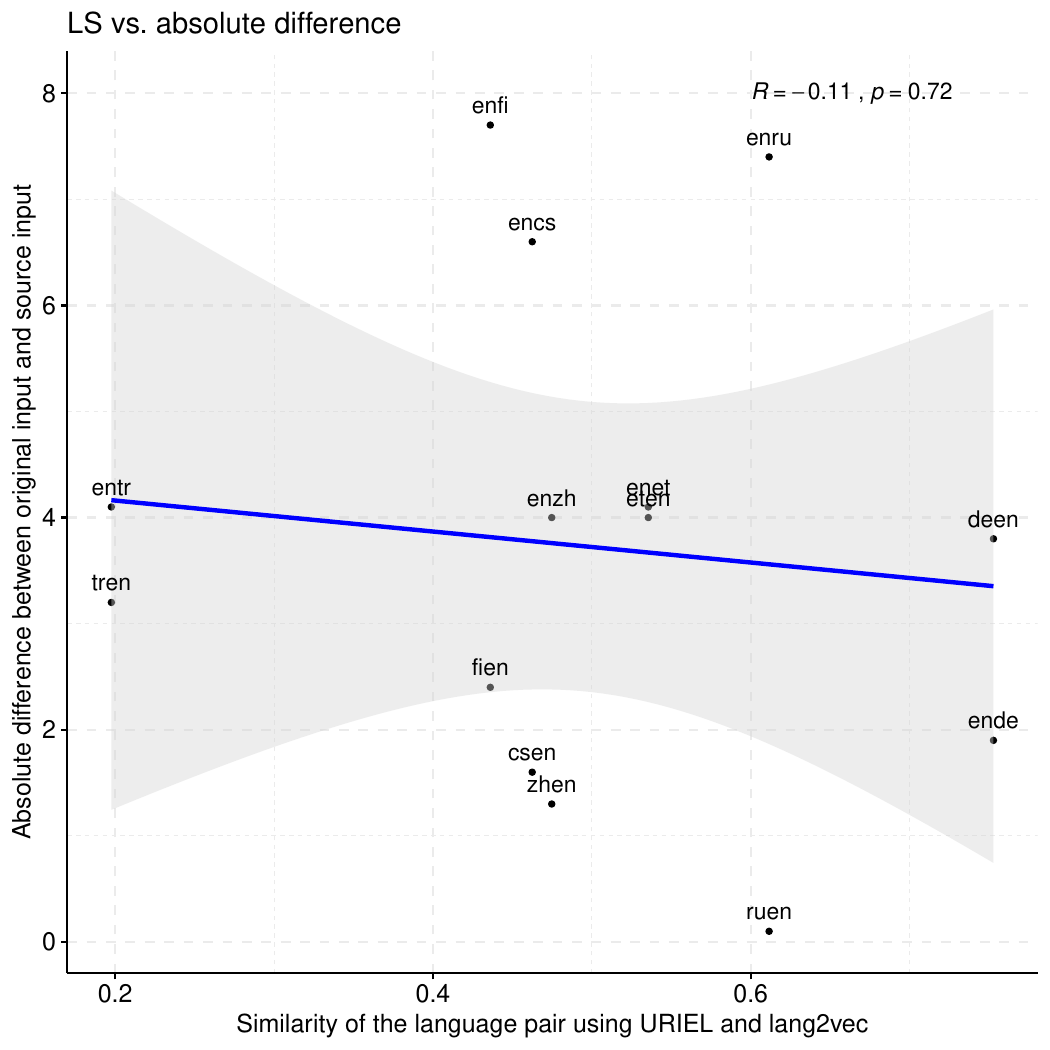}
    \caption{Pearson correlation between Linguistic Similarity for each language direction and the relative (left) and absolute (right) difference (\%) in DA score of comparing WMT input and ORG input. The languages are abbreviated into ISO 639-1 codes~\citep{byrum1999iso}.}
    \label{fig:sim.vs.abs/rel}
\end{figure*}

\autoref{fig:org.vs.abs/rel} shows the Pearson correlation and 95\% confidence region of the DA score of the best scoring system for each language direction on subset ORG against the absolute and relative difference of the DA scores of those systems between WMT input and ORG input. 
We observe an interesting trend; 
higher scoring systems tend to have lower differences in score, which indicates that translationese has less effect.
Considering either relative or absolute differences, the correlations are in both cases significant and strong ($p<0.001$, $|R|>0.75$).

\autoref{fig:sim.vs.abs/rel} shows the Pearson correlation and 95\% confidence region of the LS of a language pair (English compared to another language in our data sets) against the absolute and relative difference of the DA scores of the best system for each translation direction between WMT input and ORG input. 
Here, we see a less obvious trend, and in fact both correlations are very weak and non-significant. 
However, just as in the previous figure we can see that most of the out-of-English systems tend to have a higher relative and absolute difference than systems that translate into English. 

On a side note, we created different feature combinations from the earlier mentioned features for LS. Apart from syntactic and family tree features, phonological features are also present in URIEL.
However, other combinations did not seem to alter the LS difference score, compared to using the mentioned features in the experimental setup. 

\section{Conclusion and Future Work}\label{s:conclusions}

This paper has looked in depth at the effect of translationese in bidirectional test sets, commonly used in machine translation shared tasks,
by conducting a series of experiments on data sets for 17 translation directions in the three last editions of the news shared task from WMT.
Specifically, we have recomputed the direct assessment (DA) scores separately for the whole test set (WMT), and for the subsets whose source side contains original language (ORG) and translationese (TRS). 
Results show that using original language input lowers the DA scores, and translationese input increases the scores (RQ1), and perhaps more importantly, system rankings do change (RQ2). 
We have also investigated the degree to which these rankings change, by
measuring the correlation between the rankings with a non-parametric correlation metric that supports ties (Kendall's $\tau$).
Results show that systems do change in absolute ranking, but tend to stay more in the same cluster as they were before. 

Last, we looked at whether the effect of translationese correlates with certain characteristics of translation directions.
We did not find a correlation between the effect of translationese and the level of relatedness of the two languages involved but we did find a correlation between the effect of translationese and the translation quality attainable for translation directions (RQ3).
In other words, human evaluation for better performing systems would seem to be less affected by translationese.
Related, we observe that translation directions that contain an under-resourced language tend to obtain low DA scores.
Hence, we could say that the effect of translationese tends to be high specially when an under-resourced language is present, which could distort (inflate) the expectations in terms of translation quality for these languages.


As for future work, we plan to focus on studying what the characteristics of translationese are. I.e. what are the traits that set apart the language used in original test sets from translationese test sets.

All the code and data used in our experiments are available on GitHub\footnote{https://github.com/jjzha/translationese}. 

\bibliography{acl2019}
\bibliographystyle{acl_natbib}

\appendix


\section{Supplemental Material}
\label{sec:supplemental}

These are the supplementary tables for the paper ``The Effect of Translationese in Machine Translation Test Sets". Provided are the remaining 16 tables of each language direction. These tables are of the same structure as Table 4 in the paper.




\begin{table*}
\resizebox{\linewidth}{!}{
\centering
\begin{tabular}{lclrr|cclrr|cclrr}
\multicolumn{15}{c}{\textbf{English$\rightarrow$Chinese}}\\\\
& \# &               SYSTEM &   RAW.WMT &   Z.WMT & \# &$\uparrow\downarrow$&  SYSTEM            &     RAW.ORG &  Z.ORG& \# &$\uparrow\downarrow$       & SYSTEM                &    RAW.TRS &    Z.TRS \\\hline\hline
&   1& SogouKnowing-nmt     &      73.2 &   0.208 &   1&$-$                 & SogouKnowing-nmt   &        69.1 &     0.063   &   1&$2^{\uparrow}$       & xmunmt                &       78.2 &    0.396 \\
&    & uedin-nmt            &      72.5 &   0.178 &    &$-$                 & uedin-nmt          &        67.8 &     0.015   &    &$1^{\downarrow}$     & SogouKnowing-nmt      &       77.3 &    0.352 \\
&    & xmunmt               &      72.0 &   0.165 &    &$-$                 & xmunmt             &        66.0 &    -0.059   &    &$1^{\downarrow}$     & uedin-nmt             &       77.4 &    0.349 \\
&   4& online-B             &      69.8 &   0.065 &    &$2^{\uparrow}$      & CASICT-cons        &        63.6 &    -0.134   &    &$1^{\uparrow}$       & jhu-nmt               &       75.4 &    0.278 \\\parbox[t]{3mm}{\multirow{1}{*}{\rotatebox[origin=c]{90}{\textbf{wmt17}}}}
&    & jhu-nmt              &      69.5 &   0.056 &    &$1^{\downarrow}$    & online-B           &        64.7 &    -0.142   &    &$1^{\downarrow}$     & online-B              &       74.8 &    0.271 \\
\hline

\end{tabular}
}
   \caption{Results of the English$\rightarrow$Chinese language direction with WMT, ORG, and TRS. Systems are ordered by standardized mean DA score. If a system does not contain a rank, it indicates that it shares the same cluster as the system above it. Clusters are obtained according to Wilcoxon rank-sum test at p-level p $\leq$ 0.05. Indicated in the [$\uparrow\downarrow$] column are the changes in absolute ranking (i.e. how many positions it goes up or down).} \label{tab:enzh}
\end{table*}

\clearpage

\begin{table*}
\resizebox{\linewidth}{!}{
\centering
\begin{tabular}{lclrr|cclrr|cclrr}
\multicolumn{15}{c}{\textbf{Czech$\rightarrow$English}}\\\\
& \# &          SYSTEM &   RAW.WMT &  Z.WMT  & \#  &$\uparrow\downarrow$& SYSTEM          &     RAW.ORG &     Z.ORG & \# &$\uparrow\downarrow$  &           SYSTEM     &    RAW.TRS &    Z.TRS \\\hline\hline
&  1 & uedin-nmt       &      75.4 &   0.207 & 1   &$-$                    & uedin-nmt       &        69.6 &    -0.010 & 1  &$-$                      & uedin-nmt            &       81.1 &    0.421 \\
&  2 & jhu-pbmt        &      72.6 &   0.101 & 2   &$-$                    & jhu-pbmt        &        67.7 &    -0.073 & 2  &$-$                      & jhu-pbmt             &       77.5 &    0.275 \\\parbox[t]{3mm}{\multirow{1}{*}{\rotatebox[origin=c]{90}{\textbf{wmt16}}}}       
&  3 & online-B        &      70.8 &   0.051 & 3   &$-$                    & online-B        &        66.1 &    -0.124 &    &$-$                      & online-B             &       75.5 &    0.224 \\
&  4 & online-A        &      69.5 &   0.000 &     &$-$                    & online-A        &        64.8 &    -0.169 & 4  &$1^{\uparrow}$           & PJATK                &       75.3 &    0.197 \\
&    & PJATK           &      69.0 &  -0.024 & 5   &$-$                    & PJATK           &        62.7 &    -0.245 &    &$1^{\downarrow}$         & online-A             &       74.0 &    0.165 \\
\hline
\end{tabular}

}
\caption{Results of the Czech$\rightarrow$English language direction with WMT, ORG, and TRS. Systems are ordered by standardized mean DA score. If a system does not contain a rank, it indicates that it shares the same cluster as the system above it. Clusters are obtained according to Wilcoxon rank-sum test at p-level p $\leq$ 0.05. Indicated in the [$\uparrow\downarrow$] column are the changes in absolute ranking (i.e. how many positions it goes up or down).} \label{tab:csen}
\end{table*}

\begin{table*}
\resizebox{\linewidth}{!}{
\begin{tabular}{lclrr|cclrr|cclrr}
\multicolumn{15}{c}{\textbf{English$\rightarrow$Czech}}\\\\
& \# &  SYSTEM                  &   RAW.WMT &   Z.WMT & \# &$\uparrow\downarrow$    &  SYSTEM                &     RAW.ORG &     Z.ORG & \# &$\uparrow\downarrow$    &  SYSTEM                &    RAW.TRS &    Z.TRS \\\hline\hline
&   1&  uedin-nmt               &      62.0 &   0.308 &  1 &$-$                     &  uedin-nmt             &        56.2 &     0.126 &  1 &$-$                     &  uedin-nmt             &       69.4 &    0.544 \\
&   2&  online-B                &      59.7 &   0.240 &    &$-$                     &  online-B              &        55.1 &     0.090 &  2 &$-$                     &  online-B              &       64.8 &    0.405 \\
&   3&  limsi-factored-norm     &      55.9 &   0.111 &   3&$-$                     &  limsi-factored-norm   &        50.2 &    -0.074 &    &$-$                     &  limsi-factored-norm   &       63.5 &    0.354 \\
&    &  LIUM-FNMT               &      55.2 &   0.102 &    &$-$                     &  LIUM-FNMT             &        49.6 &    -0.076 &    &$1^{\uparrow}$          &  LIUM-NMT              &       61.8 &    0.299 \\
&    &  LIUM-NMT                &      55.2 &   0.090 &    &$-$                     &  LIUM-NMT              &        49.5 &    -0.086 &    &$1^{\downarrow}$        &  LIUM-FNMT             &       61.5 &    0.299 \\
\hline
\end{tabular}
}
   \caption{Results of the English$\rightarrow$Czech language direction with WMT, ORG, and TRS. Systems are ordered by standardized mean DA score. If a system does not contain a rank, it indicates that it shares the same cluster as the system above it. Clusters are obtained according to Wilcoxon rank-sum test at p-level p $\leq$ 0.05. Indicated in the [$\uparrow\downarrow$] column are the changes in absolute ranking (i.e. how many positions it goes up or down).} \label{tab:encs}
\end{table*}

\clearpage

\begin{table*}
\resizebox{\linewidth}{!}{
\centering
\begin{tabular}{lclrr|cclrr|cclrr}
\multicolumn{15}{c}{\textbf{Estonian$\rightarrow$English}}\\\\
& \# & SYSTEM                &   RAW.WMT &   Z.WMT & \#  &$\uparrow\downarrow$  &  SYSTEM                &     RAW.ORG &     Z.ORG & \# &$\uparrow\downarrow$   &   SYSTEM               &    RAW.TRS &    Z.TRS \\\hline\hline
&   1&     tilde-nc-nmt      &      73.3 &   0.326 &    1&$-$                   &      tilde-nc-nmt      &        69.3 &     0.208 &   1&$-$                    &      tilde-nc-nmt      &       77.3 &    0.444 \\
&   2&             NICT      &      71.1 &   0.238 &    2&$-$                   &              NICT      &        66.2 &     0.108 &   2&$-$                    &              NICT      &       75.8 &    0.366 \\
&    &      tilde-c-nmt      &      69.9 &   0.215 &     &$-$                   &       tilde-c-nmt      &        66.1 &     0.101 &    &$-$                    &       tilde-c-nmt      &       73.9 &    0.331 \\
&    &           M4t1ss      &      69.0 &   0.187 &     &$1^{\uparrow}$        &             uedin      &        65.6 &     0.060 &    &$-$                    &            M4t1ss      &       73.5 &    0.316 \\
&    &            uedin      &      69.2 &   0.186 &     &$1^{\downarrow}$      &            M4t1ss      &        64.4 &     0.058 &    &$1^{\uparrow}$         &  tilde-c-nmt-comb      &       73.0 &    0.312 \\
\hline
\end{tabular}

}
   \caption{Results of the Estonian$\rightarrow$English language direction with WMT, ORG, and TRS. Systems are ordered by standardized mean DA score. If a system does not contain a rank, it indicates that it shares the same cluster as the system above it. Clusters are obtained according to Wilcoxon rank-sum test at p-level p $\leq$ 0.05. Indicated in the [$\uparrow\downarrow$] column are the changes in absolute ranking (i.e. how many positions it goes up or down).} \label{tab:eten}
\end{table*}

\begin{table*}
\resizebox{\linewidth}{!}{
\centering
\begin{tabular}{lclrr|cclrr|cclrr}
\multicolumn{15}{c}{\textbf{English$\rightarrow$Estonian}}\\\\
& \# & SYSTEM             &   RAW.WMT &   Z.WMT & \# &$\uparrow\downarrow$      &  SYSTEM             &     RAW.ORG &     Z.ORG & \# &$\uparrow\downarrow$      &   SYSTEM            &    RAW.TRS &    Z.TRS \\\hline\hline
&   1&  tilde-nc-nmt      &      64.9 &   0.549 &   1& $-$                      &   tilde-nc-nmt      &        60.8 &     0.416 &   1& $-$                      &   tilde-nc-nmt      &       68.8 &    0.676 \\
&   2&          NICT      &      62.1 &   0.453 &    & $-$                      &           NICT      &        59.2 &     0.346 &   2& $-$                      &           NICT      &       64.9 &    0.558 \\
&    &   tilde-c-nmt      &      61.6 &   0.427 &    & $-$                      &    tilde-c-nmt      &        58.5 &     0.317 &    & $1^{\uparrow}$           &         M4t1ss      &       64.9 &    0.545 \\
&    &        M4t1ss      &      61.2 &   0.418 &    & $-$                      &         M4t1ss      &        57.5 &     0.289 &    & $1^{\downarrow}$         &    tilde-c-nmt      &       64.6 &    0.534 \\
&   5&         Aalto      &      58.6 &   0.340 &    & $-$                      &          Aalto      &        55.1 &     0.213 &    & $1^{\uparrow}$           &   HY-NMT-en-et      &       62.7 &    0.464 \\
\hline
\end{tabular}

}
\caption{Results of the English$\rightarrow$Estonian language direction with WMT, ORG, and TRS. Systems are ordered by standardized mean DA score. If a system does not contain a rank, it indicates that it shares the same cluster as the system above it. Clusters are obtained according to Wilcoxon rank-sum test at p-level p $\leq$ 0.05. Indicated in the [$\uparrow\downarrow$] column are the changes in absolute ranking (i.e. how many positions it goes up or down).} \label{tab:enet}
\end{table*}

\clearpage

\begin{table*}
\resizebox{\linewidth}{!}{
\centering
\begin{tabular}{lclrr|cclrr|cclrr}
\multicolumn{15}{c}{\textbf{Finnish$\rightarrow$English}}\\\\
& \# &       SYSTEM          &   RAW.WMT &   Z.WMT & \# & $\uparrow\downarrow$  &        SYSTEM         &     RAW.ORG &     Z.ORG & \# & $\uparrow\downarrow$   &        SYSTEM          &    RAW.TRS &    Z.TRS \\\hline\hline

&   1&     online-B          &      66.9 &   0.095 &   1& $-$  &      online-B         &        63.7 &    -0.005 &   1& $2^{\uparrow}$                          &      online-G          &       69.9 &    0.220 \\
&    &   uedin-pbmt          &      66.3 &   0.087 &    & $-$  &    uedin-pbmt         &        63.1 &    -0.034 &    & $-$                                     &    uedin-pbmt          &       69.6 &    0.210 \\
&    &     online-G          &      66.4 &   0.084 &    & $-$  &      online-G         &        62.9 &    -0.051 &    & $1^{\uparrow}$                          &       UH-opus          &       70.0 &    0.207 \\\parbox[t]{3mm}{\multirow{1}{*}{\rotatebox[origin=c]{90}{\textbf{wmt16}}}} 
&    &      UH-opus          &      65.9 &   0.065 &    & $-$  &       UH-opus         &        61.8 &    -0.078 &    & $3^{\downarrow}$                        &      online-B          &       70.1 &    0.195 \\
&   5&    PROMT-SMT          &      62.9 &  -0.037 &   5& $-$  &     PROMT-SMT         &        60.3 &    -0.136 &   5& $-$                                     &     PROMT-SMT          &       65.5 &    0.063 \\
\hline

\end{tabular}
}
   \caption{Results of the Finnish$\rightarrow$English language direction with WMT, ORG, and TRS. Systems are ordered by standardized mean DA score. If a system does not contain a rank, it indicates that it shares the same cluster as the system above it. Clusters are obtained according to Wilcoxon rank-sum test at p-level p $\leq$ 0.05. Indicated in the [$\uparrow\downarrow$] column are the changes in absolute ranking (i.e. how many positions it goes up or down).} \label{tab:fien}
\end{table*}

\begin{table*}
\resizebox{\linewidth}{!}{
\centering
\begin{tabular}{lclrr|cclrr|cclrr}
\multicolumn{15}{c}{\textbf{English$\rightarrow$Finnish}}\\\\
& \# & SYSTEM                       &   RAW.WMT &   Z.WMT & \# &$\uparrow\downarrow$    &  SYSTEM                       & RAW.ORG &     Z.ORG & \# &$\uparrow\downarrow$    &   SYSTEM                  &    RAW.TRS &    Z.TRS \\\hline\hline
&   1&                   online-B   &      59.6 &   0.378 &   1&$-$                     &                    online-B   &    54.5 &     0.209 &   1&$-$                     &                 online-B  &       65.2 &    0.561 \\
&    &                 HY-HNMT      &      57.8 &   0.305 &   2&$-$                     &                  HY-HNMT      &    51.7 &     0.096 &    &$-$                     &                  HY-HNMT  &       64.4 &    0.534 \\
&   3&                   online-G   &      51.6 &   0.090 &   3&$-$                     &                    online-G   &    48.5 &    -0.026 &   3&$2^{\uparrow}$          &       AaltoHnmtMulti      &       55.8 &    0.236 \\
&    & jhu-nmt-lattice              &      51.3 &   0.060 &   4&$-$                     &          jhu-nmt-lattice      &    46.4 &    -0.108 &    &$-$                     &          jhu-nmt-lattice  &       56.0 &    0.220 \\
&    &      AaltoHnmtMulti          &      49.3 &  -0.004 &    &$-$                     &       AaltoHnmtMulti          &    43.8 &    -0.208 &    &$2^{\downarrow}$        &                 online-G  &       54.6 &    0.201 \\
\hline

\end{tabular}

}
   \caption{Results of the English$\rightarrow$Finnish language direction with WMT, ORG, and TRS. Systems are ordered by standardized mean DA score. If a system does not contain a rank, it indicates that it shares the same cluster as the system above it. Clusters are obtained according to Wilcoxon rank-sum test at p-level p $\leq$ 0.05. Indicated in the [$\uparrow\downarrow$] column are the changes in absolute ranking (i.e. how many positions it goes up or down).} \label{tab:enfi}
\end{table*}
\clearpage

\begin{table*}
\resizebox{\linewidth}{!}{
\centering
\begin{tabular}{lclrr|cclrr|cclrr}
\multicolumn{15}{c}{\textbf{German$\rightarrow$English}}\\\\
& \# & SYSTEM                 &   RAW.WMT &   Z.WMT & \# &$\uparrow\downarrow$  &        SYSTEM           &     RAW.ORG &     Z.ORG & \#  &$\uparrow\downarrow$  &        SYSTEM           &    RAW.TRS &    Z.TRS \\\hline\hline
&   1&    uedin-nmt           &      75.8 &   0.204 &   1&$-$                   &     uedin-nmt           &        71.7 &     0.051 &    1& $-$                  &     uedin-nmt           &       79.9 &    0.357 \\
&   2&     online-A           &      72.7 &   0.095 &   2&$-$                   &      online-A           &        68.4 &    -0.046 &    2& $1^{\uparrow}$       &      online-B           &       77.2 &    0.251 \\
&    &     online-B           &      72.2 &   0.086 &    &$-$                   &      online-B           &        67.3 &    -0.079 &     & $1^{\uparrow}$       &  uedin-syntax           &       76.4 &    0.240 \\
&    & uedin-syntax           &      71.5 &   0.065 &    &$-$                   &  uedin-syntax           &        66.6 &    -0.108 &     & $1^{\uparrow}$       &           KIT           &       76.6 &    0.237 \\\parbox[t]{3mm}{\multirow{1}{*}{\rotatebox[origin=c]{90}{\textbf{wmt16}}}} 
&    &          KIT           &      71.4 &   0.062 &    &$-$                   &           KIT           &        66.2 &    -0.112 &     & $3^{\downarrow}$     &      online-A           &       77.0 &    0.235 \\
\hline
\end{tabular}
}
   \caption{Results of the German$\rightarrow$English language direction with WMT, ORG, and TRS. Systems are ordered by standardized mean DA score. If a system does not contain a rank, it indicates that it shares the same cluster as the system above it. Clusters are obtained according to Wilcoxon rank-sum test at p-level p $\leq$ 0.05. Indicated in the [$\uparrow\downarrow$] column are the changes in absolute ranking (i.e. how many positions it goes up or down).} \label{tab:deen}
\end{table*}

\begin{table*}
\resizebox{\linewidth}{!}{
\centering
\begin{tabular}{lclrr|cclrr|cclrr}
\multicolumn{15}{c}{\textbf{English$\rightarrow$German}}\\\\
& \# & SYSTEM                   &   RAW.WMT &   Z.WMT & \# &$\uparrow\downarrow$    &  SYSTEM                     &     RAW.ORG &     Z.ORG & \# &$\uparrow\downarrow$  &   SYSTEM &  RAW.TRS         &  Z.TRS \\\hline\hline  
&   1&    LMU-nmt-reranked      &      72.9 &   0.257 &   1&$-$                     &     LMU-nmt-reranked        &        68.8 &     0.101 &   1& $-$                  &     LMU-nmt-reranked        &       77.3 &    0.423 \\
&   2&               online-B   &      70.2 &   0.158 &    &$-$                     &                online-B     &        66.9 &     0.052 &    & $1^{\uparrow}$       &            uedin-nmt        &       75.9 &    0.356 \\
&    &           uedin-nmt      &      69.8 &   0.139 &    &$-$                     &            uedin-nmt        &        65.0 &    -0.036 &    & $1^{\downarrow}$     &                online-B     &       74.4 &    0.294 \\
&    &             SYSTRAN      &      68.9 &   0.092 &    &$-$                     &              SYSTRAN        &        64.4 &    -0.059 &    & $1^{\uparrow}$       &       LMU-nmt-single        &       73.3 &    0.280 \\
&    &      LMU-nmt-single      &      66.9 &   0.035 &    &$4^{\uparrow}$          &    RWTH-nmt                 &        61.9 &    -0.149 &    & $1^{\downarrow}$     &              SYSTRAN        &       73.9 &    0.256 \\
\hline
\end{tabular}
}
\caption{Results of the English$\rightarrow$German language direction with WMT, ORG, and TRS. Systems are ordered by standardized mean DA score. If a system does not contain a rank, it indicates that it shares the same cluster as the system above it. Clusters are obtained according to Wilcoxon rank-sum test at p-level p $\leq$ 0.05. Indicated in the [$\uparrow\downarrow$] column are the changes in absolute ranking (i.e. how many positions it goes up or down).} \label{tab:ende}
\end{table*}
\clearpage

\begin{table*}
\resizebox{\linewidth}{!}{
\centering
\begin{tabular}{lclrr|cclrr|cclrr}
\multicolumn{15}{c}{\textbf{Latvian$\rightarrow$English}}\\\\
& \# &                      SYSTEM &   RAW.WMT &   Z.WMT & \# &$\uparrow\downarrow$ &  SYSTEM                      &  RAW.ORG    &     Z.ORG & \# &$\uparrow\downarrow$ &  SYSTEM                     &  RAW.TRS   &    Z.TRS \\\hline\hline
&   1&                  online-B   &      76.2 &   0.266 &   1&$-$                  &                   online-B   &        75.8 &     0.266 &   1&$1^{\uparrow}$       & tilde-nc-nmt-smt            &       76.8 &    0.268 \\
&    & tilde-nc-nmt-smt            &      76.2 &   0.245 &    &$-$                  &  tilde-nc-nmt-smt            &        75.5 &     0.222 &    &$1^{\downarrow}$     &                  online-B   &       76.5 &    0.267 \\
&   3&               uedin-nmt     &      71.4 &   0.087 &   3&$1^{\uparrow}$       &   tilde-c-nmt-smt            &        69.8 &     0.043 &   3&$-$                  &               uedin-nmt     &       74.0 &    0.168 \\\parbox[t]{3mm}{\multirow{1}{*}{\rotatebox[origin=c]{90}{\textbf{wmt17}}}} 
&    &  tilde-c-nmt-smt            &      71.0 &   0.083 &    &$1^{\downarrow}$     &                uedin-nmt     &        68.8 &     0.007 &   4&$-$                  &  tilde-c-nmt-smt            &       72.1 &    0.121 \\
&   5&                  online-A   &      67.3 &  -0.039 &   5&$-$                  &                   online-A   &        64.5 &    -0.142 &    &$-$                  &                  online-A   &       70.0 &    0.062 \\
\hline
\end{tabular}
}
   \caption{Results of the Latvian$\rightarrow$English language direction with WMT, ORG, and TRS. Systems are ordered by standardized mean DA score. If a system does not contain a rank, it indicates that it shares the same cluster as the system above it. Clusters are obtained according to Wilcoxon rank-sum test at p-level p $\leq$ 0.05. Indicated in the [$\uparrow\downarrow$] column are the changes in absolute ranking (i.e. how many positions it goes up or down).} \label{tab:lven}
\end{table*}

\begin{table*}
\resizebox{\linewidth}{!}{
\centering
\begin{tabular}{lclrr|cclrr|cclrr}
\multicolumn{15}{c}{\textbf{English$\rightarrow$Latvian}}\\\\
& \# & SYSTEM                       &  RAW.WMT  &  Z.WMT  & \# &$\uparrow\downarrow$    &  SYSTEM                       &  RAW.ORG    &  Z.ORG    & \# &$\uparrow\downarrow$&   SYSTEM                      &  RAW.TRS   &    Z.TRS \\\hline\hline
&   1& tilde-nc-nmt-smt             &      54.4 &   0.196 &   1&$-$                     &  tilde-nc-nmt-smt             &        43.2 &    -0.168 &   1&$-$                 &  tilde-nc-nmt-smt             &       66.1 &    0.579 \\
&    &                   online-B   &      51.6 &   0.121 &    &$-$                     &                    online-B   &        40.6 &    -0.222 &    &$1^{\uparrow}$      &   tilde-c-nmt-smt             &       64.9 &    0.519 \\
&    &  tilde-c-nmt-smt             &      51.1 &   0.104 &    &$1^{\uparrow}$          &      limsi-factored-norm      &        41.3 &    -0.235 &    &$1^{\downarrow}$    &                    online-B   &       63.1 &    0.484 \\
&    &     limsi-factored-norm      &      50.8 &   0.075 &    &$3^{\uparrow}$          &       usfd-consensus-kit      &        40.4 &    -0.244 &    &$1^{\uparrow}$      &      usfd-consensus-qt21      &       61.0 &    0.413 \\
&    &     usfd-consensus-qt21      &      50.0 &   0.058 &    &$2^{\downarrow}$        &   tilde-c-nmt-smt             &        39.2 &    -0.255 &    &$1^{\downarrow}$    &      limsi-factored-norm      &       61.0 &    0.410 \\
\hline
\end{tabular}
}
   \caption{Results of the English$\rightarrow$Latvian language direction with WMT, ORG, and TRS. Systems are ordered by standardized mean DA score. If a system does not contain a rank, it indicates that it shares the same cluster as the system above it. Clusters are obtained according to Wilcoxon rank-sum test at p-level p $\leq$ 0.05. Indicated in the [$\uparrow\downarrow$] column are the changes in absolute ranking (i.e. how many positions it goes up or down).} \label{tab:enlv}
\end{table*}
\clearpage

\begin{table*}
\resizebox{\linewidth}{!}{
\centering
\begin{tabular}{lclrr|cclrr|cclrr}
\multicolumn{15}{c}{\textbf{Romanian$\rightarrow$English}}\\\\
& \# & SYSTEM       &  RAW.WMT  &  Z.WMT  & \# &$\uparrow\downarrow$&  SYSTEM       &     RAW.ORG &     Z.ORG & \#  &$\uparrow\downarrow$&   SYSTEM      &  RAW.TRS   &    Z.TRS \\\hline\hline
&   1&     online-B &      73.9 &   0.129 &   1&$-$&      online-B &        73.5 &     0.117 &   1&$-$              &      online-B &       74.4 &    0.140 \\
&   2&    uedin-nmt &      71.2 &   0.044 &   2&$-$&     uedin-nmt &        70.9 &     0.037 &   2&$1^{\uparrow}$   &    uedin-pbmt &       72.1 &    0.063 \\\parbox[t]{3mm}{\multirow{1}{*}{\rotatebox[origin=c]{90}{\textbf{wmt16}}}} 
&    &   uedin-pbmt &      71.0 &   0.025 &    &$-$&    uedin-pbmt &        69.9 &    -0.013 &    &$2^{\uparrow}$   &      online-A &       72.2 &    0.058 \\
&    & uedin-syntax &      69.9 &  -0.000 &    &$-$&  uedin-syntax &        68.6 &    -0.031 &    &$2^{\downarrow}$ &     uedin-nmt &       71.4 &    0.052 \\
&    &     online-A &      69.7 &  -0.012 &    &$-$&      online-A &        67.2 &    -0.082 &    &$1^{\downarrow}$ &  uedin-syntax &       71.2 &    0.030 \\
&   6&        LIMSI &      66.7 &  -0.123 &   6&$-$&         LIMSI &        63.1 &    -0.257 &    &$-$              &         LIMSI &       70.3 &    0.012 \\
&    &     jhu-pbmt &      65.7 &  -0.160 &    &$-$&      jhu-pbmt &        60.6 &    -0.306 &    &$-$              &      jhu-pbmt &       70.8 &   -0.012 \\\hline
\end{tabular}
}
   \caption{Results of the Romanian$\rightarrow$English language direction with WMT, ORG, and TRS. Systems are ordered by standardized mean DA score. If a system does not contain a rank, it indicates that it shares the same cluster as the system above it. Clusters are obtained according to Wilcoxon rank-sum test at p-level p $\leq$ 0.05. Indicated in the [$\uparrow\downarrow$] column are the changes in absolute ranking (i.e. how many positions it goes up or down).} \label{tab:roen}
\end{table*}
\clearpage

\begin{table*}
\resizebox{\linewidth}{!}{
\centering
\begin{tabular}{lclrr|cclrr|cclrr}
\multicolumn{15}{c}{\textbf{Russian$\rightarrow$English}}\\\\
& \# & SYSTEM                  &  RAW.WMT  &  Z.WMT  & \# &$\uparrow\downarrow$ &  SYSTEM                 &  RAW.ORG    &  Z.ORG    & \# &$\uparrow\downarrow$  &   SYSTEM                 &    RAW.TRS &    Z.TRS \\\hline\hline
&   1&            online-G     &      74.2 &   0.115 &   1&$4^{\uparrow}$       &     PROMT-Rule-based    &        73.0 &     0.072 &   1&$-$                   &             online-G     &       76.0 &    0.172 \\
&    &           AMU-UEDIN     &      73.3 &   0.103 &    &$1^{\downarrow}$     &             online-G    &        72.5 &     0.058 &    &$-$                   &            AMU-UEDIN     &       74.6 &    0.155 \\
&    &            online-B     &      72.8 &   0.083 &    &$1^{\downarrow}$     &            AMU-UEDIN    &        72.0 &     0.051 &    &$-$                   &             online-B     &       74.8 &    0.142 \\
&    &                 NRC     &      72.7 &   0.060 &    &$1^{\downarrow}$     &             online-B    &        70.8 &     0.025 &    &$-$                   &                  NRC     &       75.0 &    0.140 \\\parbox[t]{3mm}{\multirow{1}{*}{\rotatebox[origin=c]{90}{\textbf{wmt16}}}} 
&   5&    PROMT-Rule-based     &      72.1 &   0.044 &    &$1^{\downarrow}$     &                  NRC    &        70.3 &    -0.020 &   5&$1^{\uparrow}$        &            uedin-nmt     &       72.3 &    0.061 \\
\hline
\end{tabular}

}
   \caption{Results of the Russian$\rightarrow$English language direction with WMT, ORG, and TRS. Systems are ordered by standardized mean DA score. If a system does not contain a rank, it indicates that it shares the same cluster as the system above it. Clusters are obtained according to Wilcoxon rank-sum test at p-level p $\leq$ 0.05. Indicated in the [$\uparrow\downarrow$] column are the changes in absolute ranking (i.e. how many positions it goes up or down).} \label{tab:ruen}

\end{table*}

\begin{table*}
\resizebox{\linewidth}{!}{
\centering
\begin{tabular}{lclrr|cclrr|cclrr}
\multicolumn{15}{c}{\textbf{English$\rightarrow$Russian}}\\\\
& \# & SYSTEM                    &  RAW.WMT  &  Z.WMT  & \# &$\uparrow\downarrow$   &  SYSTEM                    &     RAW.ORG &     Z.ORG & \# &$\uparrow\downarrow$   &   SYSTEM                   &  RAW.TRS   &    Z.TRS \\\hline\hline
&   1&                online-B   &      75.4 &   0.402 &   1&$-$                    &                 online-B   &        69.6 &     0.202 &   1&$-$                    &                 online-B   &       81.2 &    0.601 \\
&   2&            uedin-nmt      &      68.2 &   0.166 &   2&$-$                    &             uedin-nmt      &        60.4 &    -0.091 &   2&$-$                    &             uedin-nmt      &       76.0 &    0.424 \\
&   3&                online-H   &      66.5 &   0.105 &    &$1^{\uparrow}$         &      PROMT-Rule-based      &        60.4 &    -0.105 &    &$-$                    &                 online-H   &       74.5 &    0.384 \\
&   4&     PROMT-Rule-based      &      65.9 &   0.080 &    &$1^{\uparrow}$         &                 online-A   &        59.2 &    -0.137 &   4&$2^{\uparrow}$         &                 online-G   &       73.6 &    0.326 \\\parbox[t]{3mm}{\multirow{1}{*}{\rotatebox[origin=c]{90}{\textbf{wmt17}}}} 
&    &                online-A   &      65.2 &   0.061 &    &$2^{\downarrow}$       &                 online-H   &        58.9 &    -0.159 &   5&$1^{\downarrow}$       &      PROMT-Rule-based      &       71.6 &    0.273 \\
\hline
\end{tabular}

}
   \caption{Results of the English$\rightarrow$Russian language direction with WMT, ORG, and TRS. Systems are ordered by standardized mean DA score. If a system does not contain a rank, it indicates that it shares the same cluster as the system above it. Clusters are obtained according to Wilcoxon rank-sum test at p-level p $\leq$ 0.05. Indicated in the [$\uparrow\downarrow$] column are the changes in absolute ranking (i.e. how many positions it goes up or down).} \label{tab:enru}

\end{table*}
\clearpage

\begin{table*}
\resizebox{\linewidth}{!}{
\centering
\begin{tabular}{lclrr|cclrr|cclrr}
\multicolumn{15}{c}{\textbf{Turkish$\rightarrow$English}}\\\\
& \# &                  SYSTEM &  RAW.WMT  &  Z.WMT  & \# &$\uparrow\downarrow$ &  SYSTEM                  &  RAW.ORG    &     Z.ORG & \# &$\uparrow\downarrow$ &   SYSTEM                 &  RAW.TRS    &    Z.TRS \\\hline\hline
&   1&     online-B            &      57.1 &   0.163 &   1&$-$                  &      online-B            &        55.5 &     0.120 &   1&$-$                  &      online-B            &       58.7 &    0.205 \\
&   2&     online-G            &      55.0 &   0.109 &   2&$-$                  &      online-G            &        53.4 &     0.057 &    &$-$                  &      online-G            &       56.4 &    0.157 \\
&   3&     online-A            &      52.2 &   0.002 &    &$-$                  &      online-A            &        51.5 &    -0.009 &   3&$-$                  &      online-A            &       52.9 &    0.012 \\\parbox[t]{3mm}{\multirow{1}{*}{\rotatebox[origin=c]{90}{\textbf{wmt16}}}} 
&   4& tbtk-syscomb            &      49.6 &  -0.077 &   4&$2^{\uparrow}$       &    dvorkanton            &        48.8 &    -0.120 &    &$1^{\uparrow}$       &     PROMT-SMT            &       51.5 &   -0.015 \\
&    &    PROMT-SMT            &      49.2 &  -0.079 &    &$1^{\downarrow}$     &  tbtk-syscomb            &        48.8 &    -0.140 &    &$1^{\downarrow}$     &  tbtk-syscomb            &       50.3 &   -0.017 \\
\hline
\end{tabular}
}
   \caption{Results of the Turkish$\rightarrow$English language direction with WMT, ORG, and TRS. Systems are ordered by standardized mean DA score. If a system does not contain a rank, it indicates that it shares the same cluster as the system above it. Clusters are obtained according to Wilcoxon rank-sum test at p-level p $\leq$ 0.05. Indicated in the [$\uparrow\downarrow$] column are the changes in absolute ranking (i.e. how many positions it goes up or down).} \label{tab:tren}
\end{table*}

\begin{table*}
\resizebox{\linewidth}{!}{
\centering
\begin{tabular}{lclrr|cclrr|cclrr}
\multicolumn{15}{c}{\textbf{English$\rightarrow$Turkish}}\\\\
& \# & SYSTEM                       &  RAW.WMT  &  Z.WMT  & \# &$\uparrow\downarrow$&  SYSTEM                       &  RAW.ORG    &     Z.ORG & \# &$\uparrow\downarrow$&                        SYSTEM &  RAW.TRS   &    Z.TRS \\\hline\hline
&   1&                   online-B   &      53.4 &   0.513 &   1&$-$&                    online-B   &        40.0 &     0.131 &   1&$-$&                    online-B   &       65.2 &    0.848 \\
&   2&               uedin-nmt      &      44.0 &   0.206 &   2&$-$&                uedin-nmt      &        32.5 &    -0.134 &   2&$-$&                uedin-nmt      &       56.5 &    0.576 \\
&   3&                   online-A   &      39.1 &   0.071 &    &$-$&                    online-A   &        28.6 &    -0.241 &    &$-$&                    online-A   &       50.3 &    0.406 \\\parbox[t]{3mm}{\multirow{1}{*}{\rotatebox[origin=c]{90}{\textbf{wmt17}}}} 
&    &                   online-G   &      35.5 &  -0.032 &    &$-$&                    online-G   &        27.5 &    -0.285 &   4&$-$&                    online-G   &       42.9 &    0.200 \\
&   5&                LIUM-NMT      &      32.2 &  -0.129 &    &$-$&                 LIUM-NMT      &        23.6 &    -0.376 &    &$-$&                 LIUM-NMT      &       41.3 &    0.132 \\
\hline
\end{tabular}
}
   \caption{Results of the English$\rightarrow$Turkish language direction with WMT, ORG, and TRS. Systems are ordered by standardized mean DA score. If a system does not contain a rank, it indicates that it shares the same cluster as the system above it. Clusters are obtained according to Wilcoxon rank-sum test at p-level p $\leq$ 0.05. Indicated in the [$\uparrow\downarrow$] column are the changes in absolute ranking (i.e. how many positions it goes up or down).} \label{tab:entr}
\end{table*}

\end{document}